\documentclass[conference]{IEEEtran}
\IEEEoverridecommandlockouts
\usepackage{cite}
\usepackage{amsmath,amssymb,amsfonts}
\usepackage{algorithmic}
\usepackage{graphicx}
\usepackage{textcomp}
\usepackage{xcolor}

\usepackage{booktabs}

\usepackage{tabularx}
\usepackage{hyperref}

\usepackage{listings}
\usepackage{tikz}

\definecolor{lightgreen}{rgb}{0.56, 0.8, 0.56}
\definecolor{red}{rgb}{0.8, 0, 0}

\def\BibTeX{{\rm B\kern-.05em{\sc i\kern-.025em b}\kern-.08em
    T\kern-.1667em\lower.7ex\hbox{E}\kern-.125emX}}
\begin{document}

\title{PerfRL: A Small Language Model Framework for Efficient Code Optimization}

\author{
\IEEEauthorblockN{Shukai Duan}
\IEEEauthorblockA{\textit{University of Southern California}\\
Los Angeles, CA, USA \\
shukaidu@usc.edu}
\and
\IEEEauthorblockN{Nikos Kanakaris}
\IEEEauthorblockA{\textit{University of Southern California} \\
Los Angeles, CA, USA \\
kanakari@usc.edu}
\and
\IEEEauthorblockN{Xiongye Xiao}
\IEEEauthorblockA{\textit{University of Southern California}\\
Los Angeles, CA, USA \\
xiongyex@usc.edu}
\and
\IEEEauthorblockN{Heng Ping}
\IEEEauthorblockA{\textit{University of Southern California}\\
Los Angeles, CA, USA \\
hping@usc.edu
}
\and
\IEEEauthorblockN{Chenyu Zhou}
\IEEEauthorblockA{\textit{University of Southern California}\\
Los Angeles, CA, USA \\
czhou691@usc.edu}
\and
\IEEEauthorblockN{Nesreen K. Ahmed}
\IEEEauthorblockA{\textit{Cisco AI Research} \\
Santa Clara, CA, USA \\
nesahmed@cisco.com}
\and
\IEEEauthorblockN{Guixiang Ma}
\IEEEauthorblockA{\textit{Intel Labs}\\
Hillsboro, OR, USA \\
guixiang.ma@intel.com}
\and
\IEEEauthorblockN{Mihai Capotă}
\IEEEauthorblockA{\textit{Intel Labs}\\
Hillsboro, OR, USA  \\
mihai.capota@intel.com}

\and
\IEEEauthorblockN{Theodore L. Willke}
\IEEEauthorblockA{\textit{DataStax}\\
Santa Clara, CA, USA \\
ted.willke@datastax.com}
\and
\IEEEauthorblockN{Shahin Nazarian}
\IEEEauthorblockA{\textit{University of Southern California}\\
Los Angeles, USA \\
shahin.nazarian@usc.edu}
\and
\IEEEauthorblockN{Paul Bogdan}
\IEEEauthorblockA{\textit{University of Southern California}\\
Los Angeles, USA \\
pbogdan@usc.edu}
}

\maketitle

\begin{abstract}
Code optimization is a challenging task requiring a substantial level of expertise from developers. Nonetheless, this level of human capacity is not sufficient considering the rapid evolution of new hardware architectures and software environments. In light of this, recent research proposes adopting machine learning and artificial intelligence techniques to automate the code optimization process.
In this paper, we introduce PerfRL, an innovative framework designed to tackle the problem of code optimization. Our framework leverages the capabilities of small language models (SLMs) and reinforcement learning (RL), facilitating a system where SLMs can assimilate feedback from their environment during the fine-tuning phase, notably through unit tests.
When benchmarked against existing models, PerfRL demonstrates superior efficiency in terms of speed and computational resource usage, attributed to its reduced need for training steps and its compatibility with SLMs. Furthermore, it substantially diminishes the risk of logical and syntactical errors.
To evaluate our framework, we conduct experiments on the PIE dataset using a lightweight large language model (i.e., CodeT5) and a new reinforcement learning algorithm, namely RRHF. For evaluation purposes, we use a list of evaluation metrics related to optimization quality and speedup. The evaluation results show that our approach achieves similar or better results compared to state-of-the-art models using shorter training times and smaller pre-trained models.
\end{abstract}


\section{Introduction}
\label{submission}



 Code optimization is the task of converting a given program to a more efficient version while retaining the same input and output~\cite{code_optimization}. It requires either the utilization of a higher level of optimization (e.g., -O3) during compilation or expert programmers to manually refactor their code to make it more optimized for certain hardware. 
Both of these tasks can be daunting with the rapid advancement in hardware, which leads to an increase in development time, bug fixing, and code optimization. Moreover, it is significantly more challenging to optimize the code for certain environments or settings. These include emerging parallel heterogeneous computing architectures, medical devices, hard-to-access communication hubs, environments requiring dynamic software updating, new software that has to run on old or new hardware and distributed mobile edge computing systems having strict power/energy budgets~\cite{gottschlich2018three}.
As a result, there has been a noticeable shift to utilizing machine learning solutions and particularly natural language processing (NLP) techniques to automate code optimization and code generation~\cite{machine_programming,gottschlich2018three, NEURIPS2024device}.

With the advent of transformer architectures~\cite{attention, 10.1007/978-3-030-79150-6_50}, language models have emerged as the default technology to perform NLP tasks.
Among other applications, prominent solutions that use language models have shown encouraging results for software-related tasks, including code generation, code optimization, defect detection, and code completion, to name a few~\cite{nijkamp2022codegen,codet5-2021, SIACHOS2025125290}.
This is so easily achievable due to the large amounts of source code available in online repositories such as GitHub, which facilitate the training process~\cite{codexglue}.
However, even if language models are capable of producing code that superficially looks correct, they are unable to check its logical and syntactical validity without any external assistance. As a result, the suggested snippets of code are not always guaranteed to work as expected~\cite{codesearchnet2019}.
To that end, the combination of language models and  Reinforcement Learning (RL) has been recently proposed in the literature~\cite{shojaee2023executionbased}. Briefly, the inclusion of RL techniques enables language models to interact with their environment and receive valuable feedback. Such feedback often comes from the execution of unit tests, where the functional correctness of a given program is confirmed.

Despite the radical changes and improvements that the combination of LLMs and RL offers to the analysis of programming languages, there has not been any significant progress as far as the problem of code optimization with SLMs is concerned.
On the one hand, the majority of the existing approaches use general-purpose datasets for fine-tuning, which in turn reduces the ability of a model to propose highly optimized versions of a given code. On the other hand, methods focusing on code optimization do not exploit RL. Thus, they are incapable of getting feedback for errors in the produced snippets of code.
Moreover, most prevailing approaches rely on LLMs whose significant demand for computation resources and high energy consumption during training and inference render them inadequate for low-cost and edge devices.

To mitigate the issues mentioned above, in this work, we propose PerfRL, a novel SLM-based framework that concentrates on the task of code optimization. Our approach fuses techniques from SLMs and RL to enable the utilization of external feedback, such as feedback from unit tests. By leveraging RL techniques, SLMs are able to receive information about the validity of the generated program. This helps the whole process to (i) \textbf{become faster} (i.e. less training is required), (ii) use a \textbf{smaller model that requires less power} and fits in low-resource devices and (iii) generate optimized code that is more likely to be \textbf{free from errors}. At the same time, the performance of the produced model \textbf{remains the same}.
Although RL and SLMs as algorithmic strategies have been proposed before, this is the first time that they are both combined and used for the code optimization problem.

\noindent \textbf{Challenges.} Broadly speaking, there are four main challenges related to the work of this paper:
(\textit{i}) How can we incorporate feedback from unit tests into the training process of an SLM model?
(\textit{ii}) How can we make an SLM perform similarly to LLMs with billions (or more) of parameters?
(\textit{iii}) How can we train such an SLM such that it generates reliable code free from errors and deals with the code optimization task?
(\textit{iv}) How can we allow edge computing, cyber-physical systems and hard-to-access devices to facilitate SLMs specialized in code optimization?

\noindent \textbf{Motivation.} When it comes to environments that require limited resources and prioritize energy efficiency, Large Language Models (LLMs) struggle to operate effectively. These conditions are particularly relevant to cyber-physical systems, edge computing, and, more recently, the Internet of Battlefield Things (IoBT). These systems and devices undergo rapid evolution and changes, both in terms of hardware and software. As such, the need for code optimization and software integration, which supports this optimization, is of vital importance.
In light of this, we argue that our proposed framework represents a groundbreaking approach to SLMs for code optimization. In fact, it democratizes AI systems research by eliminating the need to train models with an extensive number of parameters, which often leads to significant expenses.

\noindent \textbf{Contributions.} The main novel contributions of this paper are the following:
\begin{itemize}
    \item We propose an end-to-end LLM-based framework for code optimization, which is capable of incorporating feedback from unit tests into its learning process using reinforcement learning (RL) techniques.
    \item Our framework is flexible and can be used with LLMs varying in size, complexity, and number of parameters or with any RL technique.
    \item We enable smaller language models (SLMs) with fewer parameters to achieve results comparable to those of LLMs with billions of parameters.
    \item We propose a novel approach that incorporates feedback from unit tests into the fine-tuning process. This allows the model to learn to produce error-free code easily.
    \item We propose a novel framework that trains SLMs suitable for remote devices, cyber-physical systems, edge computing or devices with limited hardware resources that require low energy consumption.
    \item We empirically test our approach using the PIE dataset, which demonstrated the superiority of our approach compared to the state-of-the-art baselines.
\end{itemize}

\section{Related Work}

LLMs have demonstrated promising results as far as the task of code generation is concerned. As a result, several language models (LM) for different programming languages have been proposed in the literature~\cite{chen2021evaluating}. For instance, CodeT5~\cite{codet5-2021} is a general language model, which is pre-trained on an extended version of the CodeSearchNet dataset~\cite{codesearchnet2019}. It builds on an encoder-decoder architecture similar to T5~\cite{t5} to learn generic text representations for programming and natural languages. 

Another family of large language models for program synthesis is CODEGEN~\cite{nijkamp2022codegen,nijkamp2023codegen2}. These language models have been trained with different numbers of parameters, ranging from 350M to 16.1B. Three datasets have been used during the training period, i.e. THEPILE, BIGQUERY and BIGPYTHON.
The CODEGEN models are actually autoregressive transformers with the objective of predicting the next token
similar to traditional models for natural languages~\cite{attention}.

More recently, the work in~\cite{pie2023} proposed the PIE dataset. PIE is a subset of the CodeNet~\cite{puri2021codenet} collection of code samples. It consists of trajectories of programs, where an individual programmer starts with a slower version of a program and makes changes towards improving its performance. The authors of PIE have used it to evaluate and improve the capacity of multiple variants of models from CODEGEN family. In particular, they fine-tuned these models to suggest faster versions of a given piece of code.

\section{Problem definition}


\noindent \textbf{Code optimization.} We consider the problem of generating optimized versions of an input code. Given a set of input programs $X$, the task is to generate a set of optimized programs $\hat{X}$ $\forall x \in X$. The optimized program versions take the same input and produce the same output as their original versions. For each $x \in X$, we generate a set of candidate programs $y \in Y$, using a sampling strategy $s \in \{\text{greedy}, \text{random}\}$. Our goal is to maximize the cost function:
\begin{equation}
    \text{cost}(x,y_{best} )= \text{eq}(R,y_{best}) + \text{perf}(y_{best})
\end{equation}
where $y_{best} \in{Y}$ is the best candidate, the term $\text{eq}(R, y_{best})$ measures the ability of a generated sequence with a given input to match the output of the unit test and the term $\text{perf}(y_{best})$ measures how the performance improvement of $y_{best}$ over $x$ for unit tests.

At the same time, we also aim to maximize the probability of generating $y_{best}$ from the distribution of the input program by learning from existing training scripts.
\begin{equation}
\theta^* = \arg\max_{\theta} P(y_{\text{best}} | x; \theta)
\end{equation}

Here, $\theta^*$ represents the optimal set of model parameters. Since recent solutions to the problem rely mostly on LLMs, the main challenge of this problem is the fact that LLMs cannot naturally interact with their environment. As a result, they can potentially produce optimized versions of a given code that look correct superficially but contain either syntactical or logical errors. Additionally, existing solutions that build on LLMs and RL do not concentrate on the task of code optimization. Thus, they cannot perform sufficiently well and may suffer from hallucinations. A different challenge of this problem is the ever-increasing size of the required models. These models require a sufficient amount of computing resources, which in turn results in an increment of energy consumption.

Considering the challenges above, we propose a different approach to train an LLM model for the code generation problem. Instead of using solely an LLM or building a general-purpose LLM model with some RL steps, we design a framework targeting the task of code generation. The proposed framework can facilitate different settings and combinations of the size of the selected LM model, the complexity of the RL algorithm and the relevance of the used dataset. 

    






\noindent\textbf{Research questions.} The research questions associated with the problem of code generation are the following: (i) How can we reduce the size of the models needed to generate optimized versions of code? (ii) How can we ensure that the generated codes are reliable, produce the expected results and are free from syntactical or logical errors?

\section{Proposed Approach} \label{sec:approach}

We propose PerfRL, a reinforcement learning-based SLM framework for code performance optimization. PerfRL advances the capability of SLMs to generate optimized code that improves program runtime, while also being logically and syntactically correct. To do so, it leverages techniques from SLMs and RL. It consists of three main components: (1) fine-tuning of the SLM model, (2) sample generation and (3) reinforcement learning supervision and correction.

\begin{figure*}
    \centering
    \includegraphics[width=0.73\textwidth]{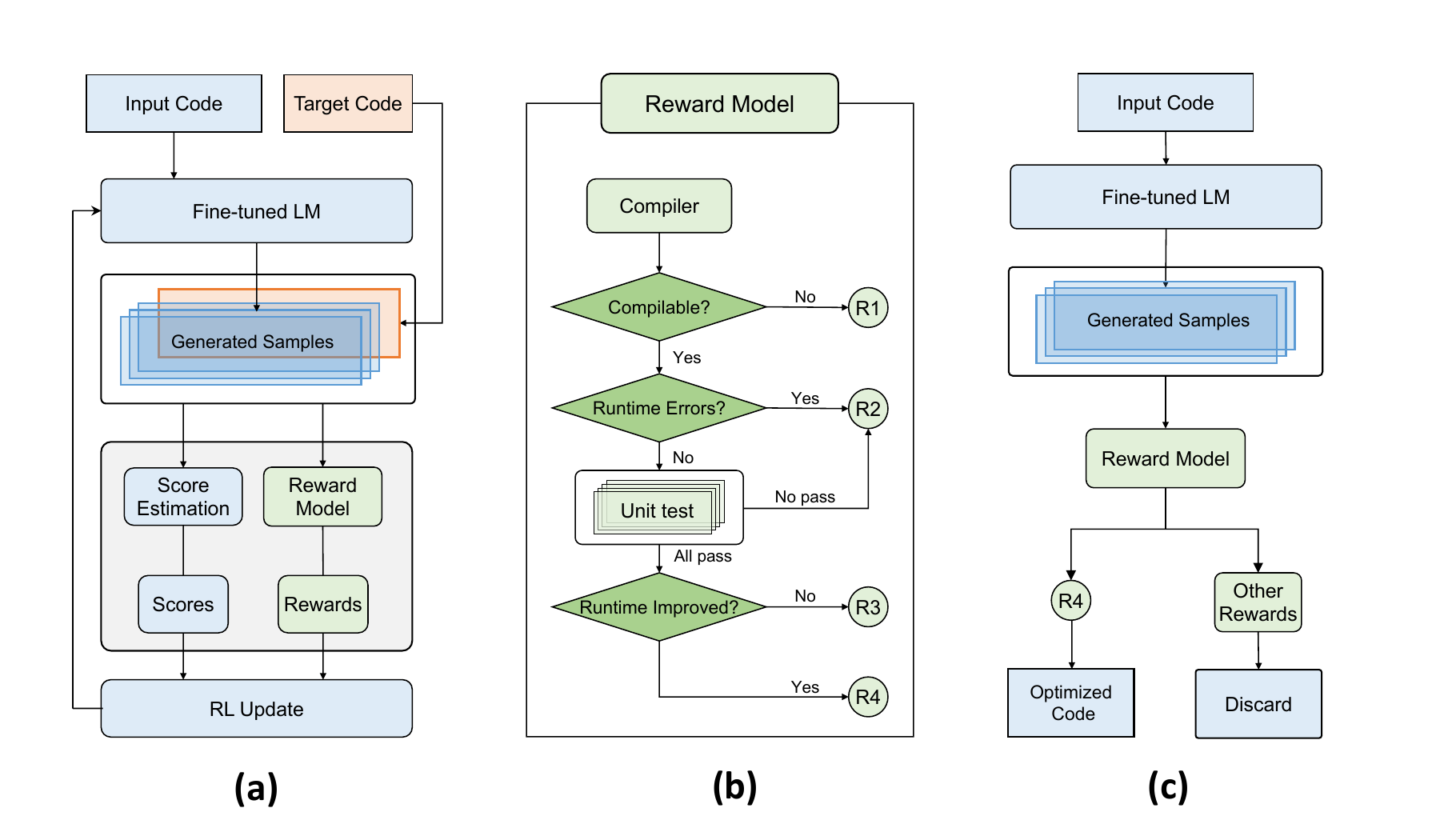}
     \caption{Overview of the PerfRL framework. \textbf{(a)~Training.} We first fine-tune an SLM model using the whole training dataset. Then, we pass the input codes into the fine-tuned LM to generate a predefined number of optimized samples for each input program. We assign a score value to each 
    sample and calculate its reward using the reward model. We utilize the score and reward values to calculate the $L_{rank}$ and $L_{tuning}$ loss values for RL. The final loss $L$ is calculated by combining the two previous loss values and is used to retrain the model. In that way, our framework incorporates feedback from unit tests into its training process. Thus, it is more likely to generate optimized code that is free from syntactical and logical errors, mitigating hallucinations. \textbf{(b)~Reward model.} Depending on the status of the code (e.g. 
    can be compiled, has a run-time error or passes all 
    the unit tests) a different reward value is given by the reward model.
    \textbf{(c)~Inference.} During inference, the final SLM is utilized to generate multiple samples of candidate source codes. These generated source codes are then evaluated using the reward model. All samples that do not receive an R4 reward are filtered out. The source codes that remain after this elimination process are considered as the optimized code.
    }
    \label{fig:approach}
\end{figure*}

Figure~\ref{fig:approach} illustrates the architecture of the proposed approach. We thoroughly describe each of the steps in this section. During the training phase, we first fine-tune an SLM using a dataset specialized in the task of code generation. Then, we utilize the fine-tuned model to generate optimized code versions for a given input code by employing different sampling strategies. We then calculate a reward value and a score for each generated code and calculate a loss value based on an RL technique. Then, during inference, we use the fine-tuned SLM model to generate an optimized version for each one of the given codes. 


\noindent \textbf{Fine-tuning of the SLM model.} The first step is to fine-tune an SLM model on a dataset that is specialized in the task of code optimization. For efficiency and simplicity, in our experiments (see Section~\ref{sec:experiments}),  we use the lightweight CodeT5 model~\cite{codet5-2021}, but our framework can operate with a variety of SLMs and LLMs, regardless of their size. In the initial paper of CodeT5, the authors use either natural language (NL) or a combination of natural and programming language (PL) as input. Particularly, they pre-train their model on tasks such as Identifier-aware denoising, Identifier Tagging, Masked Identifier Prediction and Bimodal Dual Generation.
Thus, we believe that this particular kind of models is more capable of understanding NL-PL inputs. As a result, we follow an NL-PL approach to feed the input into the CodeT5 model. Meanwhile, the authors in~\cite{pie2023} argue that a few-shot sampling strategy is beneficial for producing an optimized output. Therefore, we concatenate the action's natural language of asking the model to improve the execution performance with the input programming language and feed them into the model.

The CodeT5 model expects as input and target the preprocessed slower and faster code queries, respectively. The objective of fine-tuning is to minimize the cross-entropy loss: 

\begin{equation} \label{eq:lft}
L_{ft}(\theta) = - \frac{1}{N} \sum_{i=1}^{N} \sum_{j=1}^{V} \log(p_{i,j})
\end{equation}

where $\theta$ is the parameter of the given SLM (in our case, CodeT5), $N$ is the number of tokens, $V$ is the vocabulary set of the tokenizer, $y$ is the embedded value of the $j$-th token in the vocabulary at position $i$ in the true output sequence, and $p_{i,j}$ is the predicted probability for the $j$-th token in the vocabulary at position $i$.

\noindent \textbf{Sample Generation.} To generate candidate samples of code, we employ different sampling strategies during training, testing, and validation. More specifically, we use greedy sampling and random sampling.

Greedy sampling with beam search generates $B$ number of distinct samples for the same input. For each step of the sequence generation, the model takes the current generated sequence of tokens of sample $k$ and computes the probability distribution of the next token over the vocabulary for the corresponding sample. The top $B$ candidate vocabulary and its cumulative probability are calculated and ranked for each sample. The top  $B$ sequences among all candidates with the greatest cumulative probability are selected to repeat the process of generating the next token.
\begin{equation}
y_t^{(1)}, \ldots, y_t^{(B)} = \text{top}_B\left\{ P(y_t | y_1^{(c)}, \ldots, y_{t-1}^{(c)}, x) \right\}
\end{equation}
where $x$ is the input of the model, $Y = (y_1, \ldots, y_T), y_t \in V$ is the output tokens from the model $c \in [0,B-1]$.

The second sampling strategy is random sampling, which generates a number of distinct samples with the same input. During the generation, we set a temperature value $Tem$ in order to affect the diversity of the output. Given the probability distribution of the given logits $l$, the scaled $l'$ equals to:
\begin{equation}
l = \log P(y_t | y_1, \ldots, y_{t-1}, x)
\end{equation}
\begin{equation}
l' = \frac{l}{Tem}
\end{equation}
The probability of each token after scaling is:
\begin{equation}
p_i = \frac{exp(l'_i)}{\sum_iexp(l'_i)}
\end{equation}

Based on $p_i$, we select the top-k tokens and randomly pick one of them. We repeat such sampling steps for $t$ times until reaching the max length or end conditions.

 The next step is to generate the code samples. We observed that the model is unable to learn from most of the initially generated samples of code since they have a syntactical or logical error; as a result, the code samples do not pass the unit test. Thus, to generate samples of code during training, 
we first apply random sampling and select two candidates from independent runs. Then, we perform one greedy sampling to find the candidate with the highest probability. Finally, to ensure that at least one correct sample exists, we include the target sequence from the dataset in the list with the samples. These four samples are then fed to the model on each step.

During validation and testing (i.e. inference), we generate 4 samples using greedy sampling with beam search and return the top-2 best candidates. For a given input, we generate two candidates for evaluation. Similar to~\cite{codexglue}, we consider a sample as successful, when it has a better execution time compared to the input code. We note that existing approaches do not incorporate into their sample strategy the aforementioned heuristic, i.e. having at least one correct sample. Thus, they are more prone to learn at least one correct source code in each step.

\noindent \textbf{Reinforcement Learning.} Our RL step builds on RRHF, which is a lightweight RL framework for tuning SLMs with feedback scores. During each RL step, our objective is to maximize the probability of generating highly rewarded pieces of code by ranking and fine-tuning the SLM model. For each RL step, we sample all the data from the training set and generate the candidate outputs. Finally, for each candidate output, we compute the score, reward and loss in order to tune the model parameter~$\theta$.

\textbf{Reward.} After the code generation step, we execute each sample of code with a Python interpreter. If an error is detected, an error message is displayed.
If a piece of code does not have syntax errors, we test it for logical errors using the associated unit tests on a single core. The execution time $et_o$ for an input code $o$ is measured during the execution, assuming that there are no runtime errors within a predefined time period ($et_o \leq timeout$). In a similar fashion to the reward function proposed in~\cite{le2022coderl}, for each sample $y \in Y$, the reward $r(y)$ is calculated as follows:
\begin{equation}
r(y^{(c)}) = 
\begin{cases} 
R1 & \text{if } y^{(c)} \text{cannot be compiled} \\
R2 & \parbox[t]{5.5cm}{if $y^{(c)}$ run-time error, timeout, or failed any unit test} \\
R3 & \text{if } y^{(c)} \text{passed all unit tests} \\
R4 & \text{if } y^{(c)} \text{passed and improved run time}
\end{cases}
\end{equation}

As described in the paper of RRHF~\cite{yuan2023rrhf}, the reward values, which are assigned to the different code samples, are irrelevant to the computation of the loss, as long as a more desirable result is associated with a higher value.

\textbf{Score Function.} For a given generated code $x$, we have a candidate sequence $y^{(c)}$, where $0<c<B$. For each $y^{(c)}$, we compute the predicted score of the sequence as the sum of the log probability of each token divided by number of tokens $t$:
\begin{equation} \label{eq:score}
p^{(c)} = \frac{\sum_t \log P(y_t^{(c)} | y_1^{(c)}, \ldots, y_{t-1}^{(c)}, x)}{\lvert\lvert y_{t}^{(c)} \rvert\rvert}
\end{equation}

From the execution of the evaluation system, we obtain our reward $r^{(c)} = r(y^{(c)})$ for each $y^{(c)}$ with a given input $x$. We maximize the loss of correct responses and minimize the wrong responses by:
\begin{equation}
L_{\text{rank}} = \sum_{r^{(a)} < r^{(b)}} \max(0, p^{(a)} - p^{(b)})
\end{equation}

Since our approach is based on RRHF, we calculate the cross-entropy for the best response similar to fine-tuning: 
\begin{equation}
L_{\text{tuning}} = -\sum_t \log P(y_{best,t} | x, y_{best,<t})
\end{equation}
where $y_{best}\text{ has the greatest } r(y_i)\text{ among all } 0<i<k$. 

In that way, PerfRL can continuously enhance its output, even in cases where the best-generated code contains syntax errors or does not have the best execution time.
As we only compare the candidate outputs based on the execution time of the input program, it is highly likely that two candidate output codes will have the same reward. The loss $L_{tuning}$ selects the $y_{best}$ based on the sampling strategy. To foster the model's propensity for independent discovery of optimization strategies, we have calibrated its learning priorities. The highest priority is assigned to learning from random samples, followed by a preference for greedy samples. The target program is designated as the final learning priority.

\textbf{Loss.} We sample all the input data and calculate the loss $L$ as a combination of $L_{rank}$ and $L_{tuning}$ for the samples from the same input. 

\begin{align}
L^z &= \left( aL^z_{\text{rank}} + L^z_{\text{tuning}} \right) \\
z &\in \text{samplefrom}(X)
\end{align}

where $X$ is all the input prompts from the dataset and $a$ is a constant.

\section{Experiments} \label{sec:experiments}

\noindent \textbf{Dataset.} To fine-tune and evaluate PerfRL, we use the dataset from PIE~\cite{pie2023}. This dataset consists of approximately 40k Python files, 88k C++ files, and 3.6k Java files. PIE captures the progressive changes made by a programmer over time to improve their code. The dataset also contains (slow, fast) pairs of code written by the same programmer. We run our experiments on the subset of the dataset that concerns the Python files. The training, test and validation sets consist of approximately 36k, 1k and 2k samples respectively. Each of the samples has at least one associated unit test file that requires a specific input and has an expected output result. 
We test the accuracy of the dataset by executing all the input source code with a 5-second execution limit. We found that the $72.4\%$ of the training data, $76.4\%$ of the testing data, and $70.8\%$ of the validation data are executable. During our training, we skip all the input code that is not originally executable in the first place. As for testing and validation, we use all the data in order to compare our approach with the baseline model.

\noindent \textbf{Setup.} We run the fine-tuning and reinforcement learning steps for an instance of the CodeT5 model with $60$ million parameters and a learning rate of $2 \times 10^{-5}$.  To reduce the training time we run the whole process for $8$ RL steps for $3$ epochs. The aforementioned RL steps and epochs have been determined by performing hyperparameter tuning.
For the training of the model, we use an NVIDIA A100 GPU graphics card with 40GB of RAM on an Ubuntu 20.04 server with 50 cores. We decided to run our experiments on real hardware, as opposed to simulators such as gem5. This enables us to collect real-world evaluation results.
Our model is trained within approximately 30 hours. Each epoch computes (i) the score of the generated sequence (see Equation~\ref{eq:score}) and (ii) the loss for all the samples of the dataset from the same input data. During the training phase, we set the temperature to 1 and top\_k to 50 for random sampling. The source code and data to reproduce the experiments can be found in the anonymized repository of the project\footnote{\url{https://anonymous.4open.science/r/Perf_LLM-6B27/}}.

As already mentioned, prior to running the reinforcement learning step, we fine-tune the CodeT5 model with a one-shot learning setting. That is to say, we feed each sample of the dataset into the model once and compute the loss as described in Equation~\ref{eq:lft}. We set the learning rate to $5 \times 10^{-5}$ and batch size to $32$. The results are available in Table~\ref{tab:results}.

\begin{table*}[t]
\caption{Evaluation results of PerfRL and baseline models. \textbf{Bold} font indicates the best score on each column. The asterisk (*) indicates the baseline model. The first block shows the results as reported in~\cite{pie2023}.}
\begin{center}
\begin{scriptsize}
\begin{tabularx}{0.81\textwidth}{lccccccc}
\toprule
\textbf{METHOD} & \textbf{Model Size} & \textbf{Sample strategy} & \textbf{\%OPT} $\uparrow$ & \textbf{SP} $\uparrow$ & \textbf{RTR} $\uparrow$ & \textbf{Energy(kJ)} $\downarrow$ & \textbf{Power(W)} $\downarrow$ \\
\midrule
CODEGEN-16B & 16B & greedy and 1-shot & 2.2 & 1.55 & 35.48 &  311040 & 1200\\
CODEGEN-2B & 2B & greedy & 8.2 & 2.32 & 56.90  & 51840 & 600\\
CODEGEN-16B & 16B & greedy & \textbf{14.6} & 1.69 & 40.83  & 311040 & 1200\\
CODEX & - & greedy & 14.3 & 2.7 & 62.96 & - & -\\
\midrule
CodeT5 (Before RL) & 60M & greedy and 0-shot & 0 & 0 & 0 & - & \textbf{250}\\
CodeT5 (24 Fine-tuning epochs)*& 60M & greedy and 0-shot & 0.5 & 2.27 & 55.95 & \textbf{27000} & \textbf{250}\\
PerfRL (8 RL steps, our approach) & 60M & greedy and 0-shot & 2.8 & \textbf{4.93} & \textbf{79.72} & \textbf{27000} & \textbf{250}\\
\bottomrule
\label{tab:results}
\end{tabularx}
\end{scriptsize}
\end{center}
\vskip -0.28in
\end{table*}

During the validation process, we use greedy sampling with a beam search of size 4 for the input code to generate candidate samples. From these 4 samples, we choose the top 2 ones with the greatest accumulated probability. Then, we set a reward value to the greedy sampling round using the reward function $r$. The validated compilation rate measures the number of rounds that pass the compilation over the total number of input codes. The validated pass rate measures the number of rounds that pass the execution of all the unit tests over the total number of input codes. The validated optimization rate measures the number of rounds that pass the execution and have a better execution time in a single core over the total number of input codes.

We also test our approach on $1000$ samples. In the entire process, we run the test before the RL framework to test the current performance of the fine-tuned model and after the RL framework in order to show the contribution of RL.

For our experiments, we set the reward values for the reward function $r$ to $R1=0$, $R2=1$, $R3=1.3$ and $R4=2$. As already mentioned, the actual reward values do not affect the performance of the RL step, as long as we assign a higher value to a much preferable sample.

\noindent \textbf{Baselines.}
We use a fine-tuned version of the CodeT5 model with $24$ epochs as our main baseline. Furthermore, we compare our results with those of the original PIE paper, where models from the family of CodeGen and Codex are used for their experiments.

\noindent \textbf{Evaluation metrics.}
To evaluate our approach we use the following evaluation metrics:
\begin{itemize}
    \item \textbf{Percent Optimized per  (\%OPT)}: The ratio of samples on the test set that are improved by a given method.
    \item \textbf{Speedup (SP)}: The actual (absolute) improvement in execution time $SP(o, n) = (\frac{o}{n})$, where $o$ and $n$ are the old and new execution times, respectively.
    \item \textbf{Runtime Reduction (RTR)}: The normalized improvement in execution time among the programs that have a decrease in runtime and are syntactically and logically correct, $RTR(o, n)=(\frac{o-n}{o} \times 100)$. We mention that the average RTR is reported over the test set.
    \item \textbf{Energy}: The total energy estimation for model training.
    \item \textbf{Power}: The total maximum power for the GPU during training.
    
\end{itemize}

\noindent To measure the execution time of each generated program, we calculate the cumulative execution time using all the unit tests.
We run each experiment three times and measure the mean time on each set of unit tests in order to ensure that the reduction of the execution time is not random.

\begin{figure}[ht]
  \centering
  \begin{minipage}{0.26\textwidth}
    \centering
    \includegraphics[width=\textwidth]{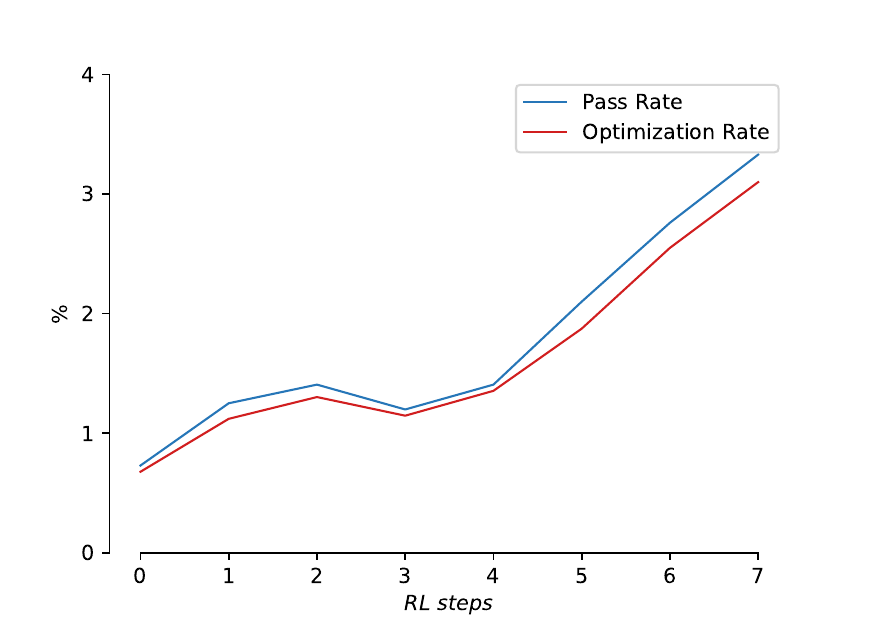}
  \end{minipage}
  \begin{minipage}{0.26\textwidth}
    \centering
    \includegraphics[width=\textwidth]{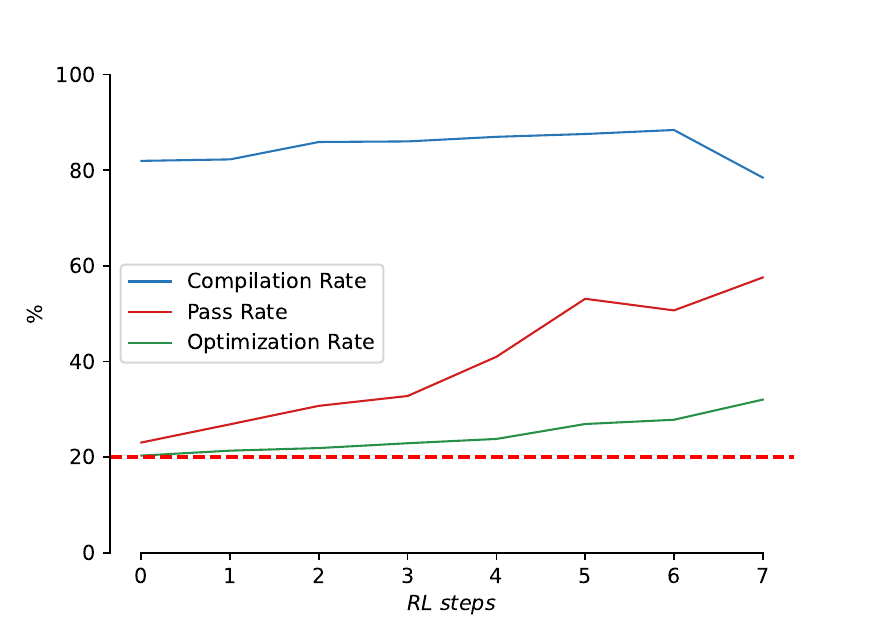}
  \end{minipage}
\caption{\textbf{(left)} Pass rate and optimization rate on validation data over RL steps on the fine-tuned CodeT5 model.
\textbf{(right)} Compilation, pass, and optimization rate per RL step for the generated programs using the fine-tuned CodeT5 model. All the rates are calculated by the number of compiled, passed, or optimized programs over the total number of the generated programs.}
\label{fig:evaluation_figure}
\vspace{-6mm}
\end{figure}

\noindent \textbf{Evaluation Results.} Table~\ref{tab:results} presents the mean \%OPT, SP, RTR, Energy and Power scores of the proposed method as well as those of the benchmark models.
Although PerfRL operates with a model that has fewer parameters, it achieves performance that is equal to or even surpasses that of its counterparts. Specifically, it significantly outperforms others in terms of the RTR and SP scores with minimum energy consumption among all models. 
Furthermore, as illustrated in the left part of Figure~\ref{fig:evaluation_figure}, both the pass and optimization rate increase proportional to the RL steps. Hence, we conclude that the utilization of an RL step enables smaller models to learn easily how to optimize source code in a less resource and time-consuming manner. More specifically, we train our model for 30 hours as opposed to the baseline CodeGen which is trained for approximately 1 or 3 days and requires a stronger machine (i.e. 2 $\times$ NVIDIA A6000 for CodeGen-2B for 1 day and 4 $\times$ NVIDIA A6000 for CodeGen-16B for 3 days).
Since the energy consumption is half of the total energy of CODEGEN-2B and 15 times less than CODEGEN-16B, we conclude that RL allows models to be trained more efficiently with less energy without the need for powerful machines.

In order to determine whether our RL step is able to generate good candidates by random or greedy sampling during the training, we measure the compilation, pass, and optimization rate as shown in the right part of Figure~\ref{fig:evaluation_figure}.
Considering our sampling strategy that concatenates the target program with the generated samples, we empirically observe that the threshold for these three values is approximately $20\%$ (red dashed line). Since these three rates are always above the threshold, they denote that our model generates meaningful candidates by itself other than purely based on the target program.

\noindent \textbf{Discussion and ablation study.} Our framework shows that an RL-based strategy for fine-tuning is able to tune a model more effectively compared to simple fine-tuning. Especially, for source code optimization, it is crucial for the model to have the ability to explore the search space by itself. However, codes that are functionally equivalent are hard to learn if their semantics are drastically different for LLM tokenizers to understand the structure of the code on limited data.
Our approach encourages SLMs and LLMs to make minor modifications to the input source code. However, we believe ML techniques with structure information of code can improve the ability of LLMs to understand the complicated structure difference between two source codes.
Demonstrating our smaller model's ability to compare with the larger model, such as CODEGEN-16B and CODEX and achieving better $SP$ and $RTR$ scores, serves as solid evidence of compatibility with both small and large models. Furthermore, our comparison of the CodeT5 model, using identical training steps, yielded better results than solely performing fine-tuning.

\section{Conclusion}
In this paper, we propose a novel framework for the task of code optimization,  called PerfRL. Our framework combines techniques from SLMs and RL, and it allows language models to take into consideration feedback from unit tests during their fine-tuning process. To demonstrate the applicability of our framework, we fine-tuned the CodeT5 model on the PIE dataset. We benchmark the proposed approach against a list of baseline models that rely solely on simple fine-tuning, while ignoring the logical and syntactical correctness of the generated code. The evaluation results demonstrate that by adopting our framework, one can reach state-of-art performance using an SLM with fewer parameters, which in turn results in lower energy consumption, something that is critical on edge computing devices and low-resource environments.

\bibliographystyle{IEEEtran}
\bibliography{IEEEabrv,main}

\end{document}